\documentclass[conference]{IEEEtran}
\IEEEoverridecommandlockouts

\usepackage{cite}
\usepackage{amsmath,amssymb,amsfonts}
\usepackage{algorithmic}
\usepackage{graphicx}
\usepackage{textcomp}
\usepackage{xcolor}
\def\BibTeX{{\rm B\kern-.05em{\sc i\kern-.025em b}\kern-.08em
    T\kern-.1667em\lower.7ex\hbox{E}\kern-.125emX}}

    \usepackage[colorinlistoftodos]{todonotes}

\usepackage[T1]{fontenc}
\usepackage{url}
\usepackage{multirow}
\usepackage{flushend}
\usepackage{caption}
\usepackage{subcaption}

\begin{document}

\title{Real-Time Onboard Object Detection for Augmented Reality: Enhancing Head-Mounted Display with YOLOv8\\
 \thanks{The study has been supported by funding provided through an unrestricted gift by Meta.}
 }


\author{
    \IEEEauthorblockN{
        Mikołaj Łysakowski\IEEEauthorrefmark{1}, Kamil Żywanowski\IEEEauthorrefmark{1},  Adam Banaszczyk\IEEEauthorrefmark{1}, Michał R. Nowicki\IEEEauthorrefmark{1}\IEEEauthorrefmark{2},\\ 
        Piotr Skrzypczyński\IEEEauthorrefmark{1}\IEEEauthorrefmark{2}, Sławomir K. Tadeja\IEEEauthorrefmark{3}
    }
    \IEEEauthorblockA{\IEEEauthorrefmark{1} Pozna\'n University of Technology, Centre for Artificial Intelligence and Cybersecurity}
    \IEEEauthorblockA{\IEEEauthorrefmark{2} Pozna\'n University of Technology, Institute of Robotics and Machine Intelligence}
    \IEEEauthorblockA{\IEEEauthorrefmark{3} University of Cambridge, Department of Engineering, Institute for Manufacturing}
}


\maketitle

\begin{abstract}
This paper introduces a software architecture for real-time object detection using machine learning (ML) in an augmented reality (AR) environment. Our approach uses the recent state-of-the-art \textit{YOLOv8} network that runs onboard on the \textit{Microsoft HoloLens 2} head-mounted display (HMD). 
The primary motivation behind this research is to enable the application of advanced ML models for enhanced perception and situational awareness with a wearable, hands-free AR platform. 
We show the image processing pipeline for the YOLOv8 model and the techniques used to make it real-time on the resource-limited edge computing platform of the headset. The experimental results demonstrate that our solution achieves real-time processing without needing offloading tasks to the cloud or any other external servers while retaining satisfactory accuracy regarding the usual mAP metric and measured qualitative performance.  
\end{abstract}


\section{Introduction}
\begin{figure}[htbp!]
	\centering
	    \includegraphics[width=\columnwidth]{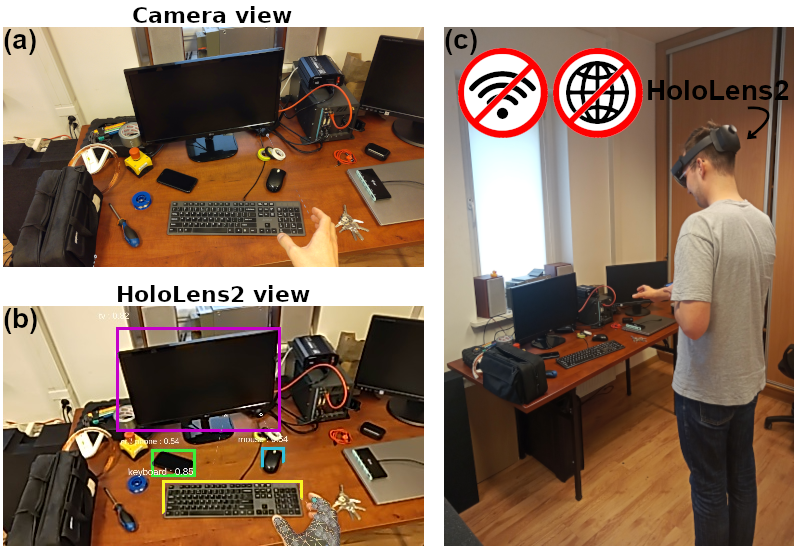}
	\caption{The proposed onboard object detection with YOLOv8 offers real-time onboard object detection enhancing HoloLens 2 capabilities without a common requirement of WiFi or Internet access to perform server-based computations.}
	\label{fig:catchy}
\end{figure}

Augmented Reality (AR) technology 
belonging to the class of \textit{immersive technologies} offers an ability to blend digital artifacts and the physical environment by superimposing digital content in the user's field of view (FoV) \cite{Caudell1992AR} (Fig.~\ref{fig:catchy}).

Presently, popular AR applications running on mobile devices, such as smartphones or tablets,   
can be further enhanced with machine learning (ML). 
Thanks to such an approach, we can include vision-based features for object detection and tracking on video and imagery data \cite{Mambu2019MobileARandML}. 

However, mobile AR solutions have significant limitations, such as a relatively small FoV confined by screen canvas or needing hand control \cite{Tadeja2022MARS}. 
The latter narrows down potential scenarios where we can successfully deploy AR, such as manual assembly \cite{Tadeja2022ARQ}, device repair task \cite{Bozzi2023Print3D}, or the use of AR enhancers by older adults \cite{Seifert2021elderly}. 
In such cases, the user's ability to not only freely move hands but promptly shift the unconstrained FoV or the body posture 
is crucial for safety concerns and task completion \cite{Syberfeldt2016TypesOfAR, Bozzi2023Print3D}.

These caveats are circumvented by the alternative technology of wearable smart head-mounted display (HMD) \cite{Tadeja2022ARQ}. The AR headsets, such as widely-considered state-of-the-art \textit{Microsoft HoloLens 2} (HL2) \cite{Hololens2Hardware}, offer a hands-free AR experience 
\cite{Tadeja2022ARQ, Bozzi2023Print3D}. Unfortunately, HL2 and other similar headsets do not offer a satisfactory level of support for ML-based processing  
that could enhance the user's ability to interact with the environment \cite{Latif2022HumanCentricAV,Seelinger2022},
Thus, having onboard, real-time ML models running in the headset's edge computing platform is crucial for developing new AR application areas.

We address the problem of real-time object detection on the HL2, including the most recent \textit{You Only Look Once} \cite{Redmon2016YOLO} YOLOv8 framework. 
We focus on defining the steps necessary to achieve a desired frame rate of image processing with the onboard ML model while identifying the constraints of the HL2 computing platform. Overcoming these limitations enables using widely-available ML algorithms on headsets. 
We also believe that AR developers can use our work on YOLOv8 for HL2 to create new applications extending the current use cases of this headset.

Our contribution can be summarized as a unique, easy-to-replicate, real-time, onboard object detection pipeline on the HL2 headset. 
Following the open-science principle, our code and complete guide on how to run the most recent YOLOv8 network architecture in the resource-limited hardware environment of this HMD is freely available as a GitHub\footnote{\url{https://github.com/kolaszko/hl2_detection}} repository. 
Furthermore, as a byproduct of our work, we also developed a commented list of limitations of the HL2 as a computing platform for ML. These should be tackled first to broaden the development of modern ML-based applications on this widely used among AR community device \cite{Tadeja2022ARQ, Bozzi2023Print3D}.

\section{Related work}
The usage of object detection on AR headsets is an item of past and current research explorations \cite{hl2_rcnn, tomato, blind, car}. For example, in~\cite{hl2_rcnn}, the authors explore the possibility of using a no longer state-of-the-art object detection approach with a two-stage network to detect and track objects while offloading the processing to the high-end server. We argue that while on the server side, we are not constrained by the computational capabilities of HL2, we cannot use the headset to carry out object-detection tasks without local WiFi or a fast wireless broadband communication (e.g., LTE connection). Off-board processing limits the HL2 headset to only a frame-capturing camera and head-mounted display output than a standalone solution for object detection and tracking. 
Presently, the Microsoft Azure Custom Vision library~\cite{azure} offers a complete high-level solution when considering off-board computations.

An example application is tomato picking, where farmers rely on inexperienced, part-time workers to harvest the fruits with desired ripeness and blemishes~\cite{tomato}. 
It negatively impacts harvest quality and work efficiency as people need to learn the task.
The presented work shows a complete AR and ML-based solution to this problem that requires powerful servers needing a steady, reliable LTE connection, which might not be available in the field.
Another example concerns a system for supporting visually-impaired people, utilizing object detection in images to provide information about objects in the surroundings using an audio interface to the user~\cite{blind}. 
The application directs users to the desired objects in the scene based on audio communication.
This system uses an old YOLOv2 model that is offloaded to the server, thus making it vulnerable to LTE connection stability outdoors, hampering its ability to efficiently determine desired objects' locations.
Similarly, in~\cite{car}, we are introduced to the concept of using HL2 as a tool to increase drivers' road condition awareness.
The authors explore the idea of a system that can focus users' attention on incoming vehicles and support lane detection while extending the view with vehicle speed.
In such a context, on-time, local processing capabilities are crucial to ensure the proper operation of the complete solution.
These examples are only a subset of the existing works concerned with object detection and tracking in headset-based AR that rely on server-side computation~\cite{yolo_students, yolo_server}, summarized in a recent review paper~\cite{yolo_review}.

The alternative to server-side processing includes simpler algorithms that run in real-time on AR headsets. 
Such an approach is proposed in~\cite{cracks}, where a custom-made processing pipeline is used to detect cracks in the constructions using only the onboard computation capabilities of HL2.
Designing custom ML pipelines offers the desired object detection accuracy but requires expert knowledge and is time-consuming, 
Hence, another approach is to apply a readily-available software framework, like \textit{Vuforia}~\cite{vuforia} or \textit{easyAR}~\cite{easyar}.
These solutions, however, require a 3D CAD model of the considered objects, which in practice limits detection capabilities to an object instance from a class of rigid objects. 

As a middle ground, it is possible to combine server-side processing with on-device edge processing. Still, the resulting solution has to properly synchronize the processing on both ends, raising even more issues~\cite{mixed_server_device}.

Consequently, we propose a pipeline that allows anybody to achieve real-time object detection and tracking capabilities using the state-of-the-art YOLOv8 network onboard HL2 without needing to be connected to any server. 
Furthermore, to achieve our real-time performance, we put a hard limit of 100 ms on end-to-end processing from an image capturing to data visualization as greater latency reduces an immersive experience to users~\cite{latency1,latency2}.
To that end, our approach offers new advantages to the ones already provided by a standalone AR headset such as HL2 \cite{Syberfeldt2016TypesOfAR}.

\section{Hardware and Software Frameworks}
\begin{figure}[!b]
	\centering
	    \includegraphics[width=0.9\columnwidth]{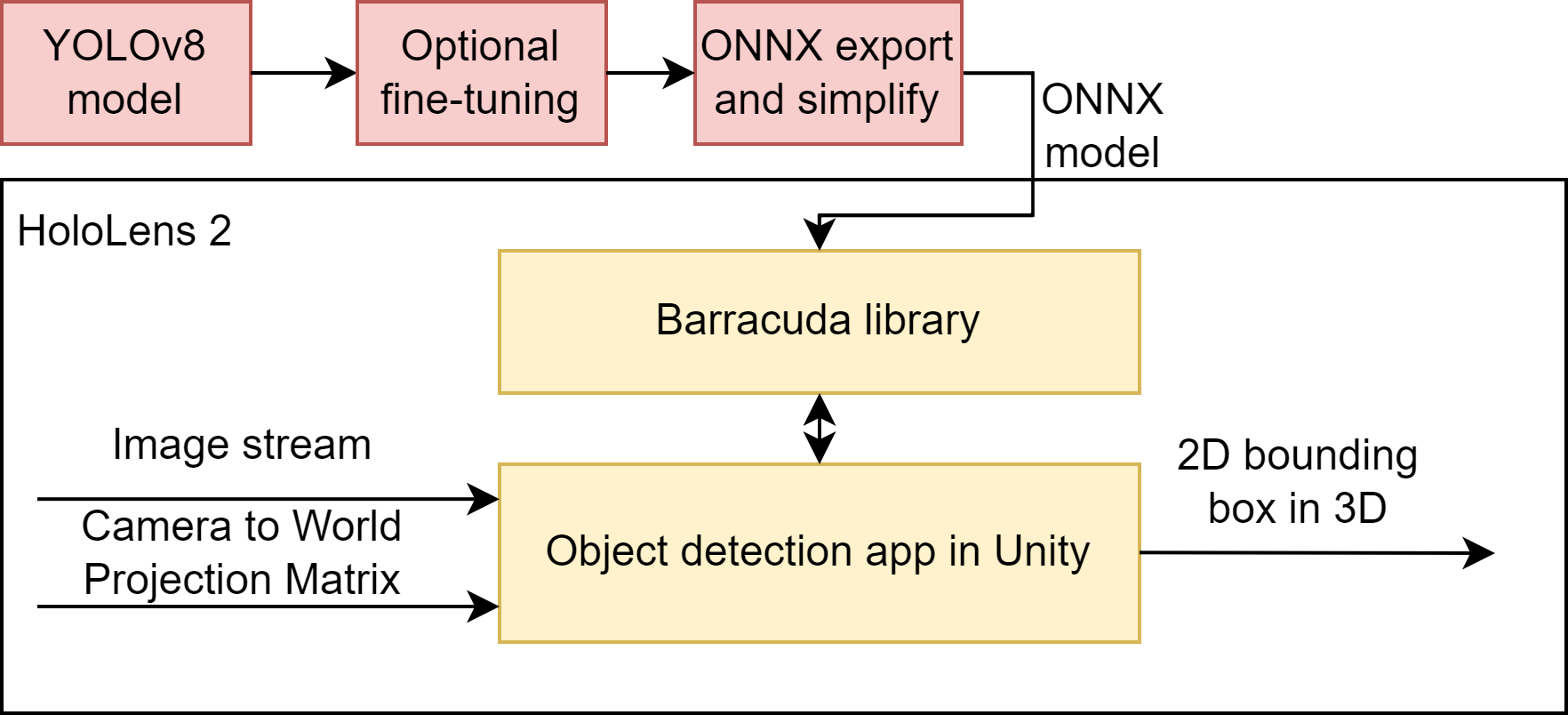}
	\caption{The relation between the used hardware and software frameworks. Notice the clear distinction between the offline phase performed outside the AR device (red) and the online operation on HL2 (yellow).}
	\label{fig:libs}
\end{figure}

The general, high-level processing idea is presented in Fig.~\ref{fig:libs}. We start by preparing the YOLOv8 neural network models for HL2. These models can be optionally retrained (fine-tuned) to include different detection classes.
The next step involves exporting the model to the \textit{Open Neural Network Exchange} (ONNX)~\cite{onnx} format. The model is then used by the \textit{Barracuda} \cite{Barracuda} library in the \textit{Unity}~\cite{Unity} engine to perform object detection on HL2 and to provide visualization of the detected objects. We decided to use the Unity platform as it is among the most widely used software framework in AR \cite{Tadeja2022ARQ, Bozzi2023Print3D} and VR (virtual reality) research \cite{edgemetaverse}. We will introduce the used frameworks in more detail in the following sections. 

\begin{figure*}[ht!]
	\centering
	    \includegraphics[width=0.8\textwidth]{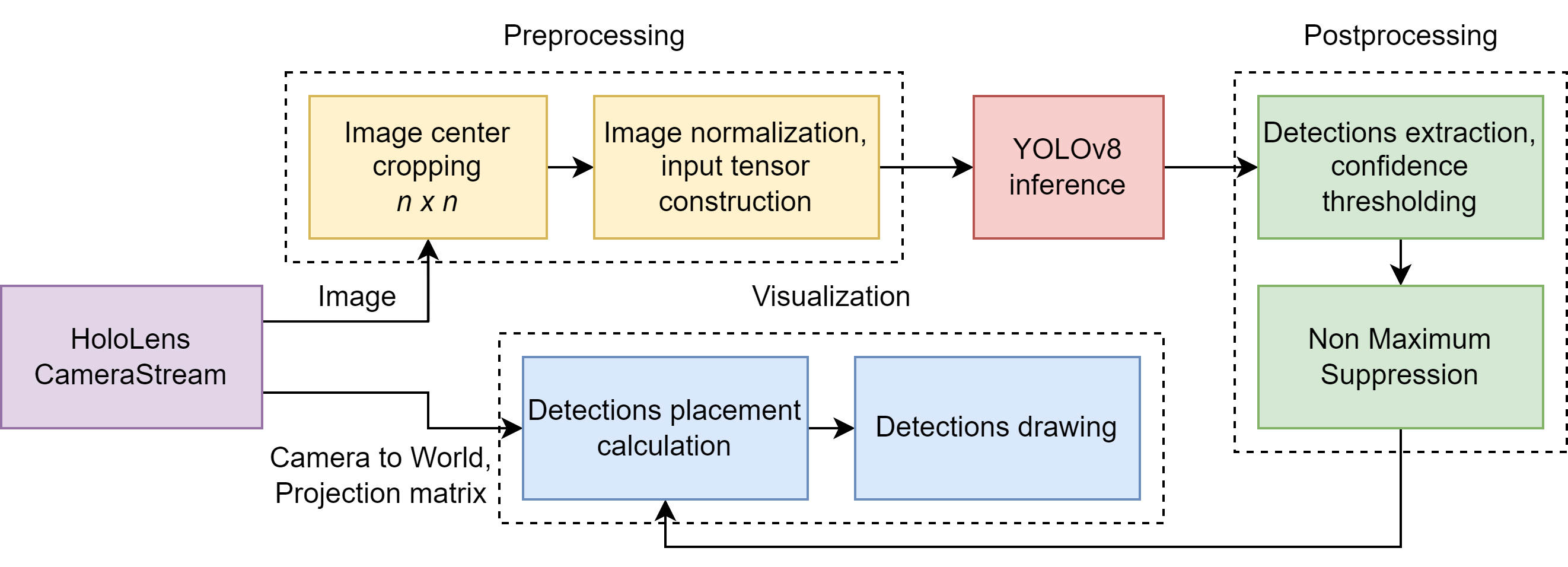}
	\caption{The image processing steps performed on the HL2 to achieve the object detection within the user's FoV.}
	\label{fig:algorithm}
\end{figure*}

\subsection{Augmented Reality Equipment: Microsoft HoloLens 2}
To showcase the possibility of optimizing an ML model performance on an HMD treated as an edge computing platform, we decided to use the \textit{HoloLens 2} \cite{Hololens2Hardware}. 
As a computing platform, this headset is equipped with Snapdragon 850, a high-performance 64-bit ARM LTE system on a chip designed by Qualcomm, and an Adreno 630 graphics processing unit (GPU). 
The headset also contains a second-generation custom-built holographic processing unit (HPU) for computations related to sensor information, core algorithm accelerators, and compute nodes enabling onboard image processing without using Snapdragon's resources. 

HL2 perception system consists of four grayscale cameras used in simultaneous localization and mapping (SLAM), two infrared (IR) cameras for the built-in eye tracking, a time-of-flight depth sensor used in hand tracking and spatial mapping, an inertial measurement unit (IMU) sensors, and frontal RGB camera.
Our work uses a world-facing RGB camera mounted above the user's eyes in the center of the front headset panel. 

All these features and their widespread application in AR-related research \cite{Tadeja2022ARQ, Bozzi2023Print3D}, made the HL2 a best-suited candidate for deploying and testing our real-time, onboard object detection with YOLOv8 network architecture.

\subsection{Object detection: You Only Look Once (YOLO) network}

Object detection is an active research topic with multiple scientific and real-world applications, as previously summarized \cite{objIn20years,modernObjDet,PerfMetr}. The YOLO family~\cite{Redmon2016YOLO} is commonly used when we need robust object detection capabilities~\cite{modernObjDet}. YOLO model is based on single-stage object detection with different backbone sizes that can be chosen based on the available processing power \cite{Redmon2016YOLO}. Real-time object detection on constrained devices can be powered by YOLO Tiny or YOLO Nano, the smallest models in the YOLO family, which however, limits the performance \cite{Roszyk2022}.

Despite offering unrivaled results, this approach is still actively explored~\cite{objIn20years}, with research focusing on achieving the best possible score (i.e., mean average precision, mAP) on the available datasets while doing it with the least amount of network parameters~\cite{PerfMetr}. 
Currently, from the YOLO family \cite{Redmon2016YOLO}, the YOLOv8 network architecture gives the best results. Hence, we will focus on this version in our demo application.




\subsection{Barracuda library for ML inference}
The neural network part of the detection pipeline is based on the Barracuda~\cite{Barracuda} library. It is an open-source library developed by Unity \cite{Unity} for utilizing neural networks in the game engine. It supports the most common deep learning layers and provides GPU and CPU inference engines. Cross-framework support for different machine learning libraries is ensured by using an ONNX format to load pretrained neural networks. 
It enables interoperability between different ML frameworks, providing a standard set of operations used in deep learning.

\subsection{Model preparation}
\label{model_prep}
Each model used in the online operation can be prepared using the same pipeline. We export each model from \textit{PyTorch} \cite{PyTorch} serialized \verb|.pt| file to the ONNX format. 
Since the current Barracuda version supports ONNX deep learning operations (opset) up to version 9, exporting models with the proper opset flag is crucial. Apart from the export, it is also possible to reduce the model with the ONNX simplification tool. The operation merges redundant operators using constant folding, consequently speeding up inference.

We successfully tested exporting and deploying the publicly available original YOLOv8 object detection models. Moreover, we can train the YOLOv8 for any custom class with sufficient data while following the guidelines for model fine-tuning to custom datasets.

\section{Object detection pipeline on HL2}


We present the universal pipeline for onboard, real-time YOLO-based object detection for HL2. The processing pipeline used in our evaluation is presented in Fig.~\ref{fig:algorithm}.

The processing starts by an image acquisition done with \verb|HoloLensCameraStream|~\cite{Hololenscamerastream} package. This plugin enables users to collect RGB camera images in all HL2-supported resolutions and frame rates, along with the current camera position in the world coordinate system. The package provides the functionality essential to calculate a projection from pixel coordinates into 3D world space using extrinsic and projection matrices. The image, current camera-to-world matrix, and projection matrix are constantly updated in a separate thread whenever new data is available.

Next, we perform the initial image preprocessing step, which consists of cropping an $n\times n$ image out of the center of the acquired camera image, where $n$ is the size of the neural network's input, and all pixel values are normalized to $[0;1]$ range. Since YOLOv8 accepts a square input, we have omitted using a rectangular image, simplifying the pre and postprocessing. Finally, the input tensor of size $(1, n, n, 3)$ is created. The image is then passed to the module for image inference with a YOLOv8-based model. The neural network structure and weights are loaded to \verb|Unity.Barracuda.Model| using an ONNX file distributed as an asset inside the application. The procedure of model preparation is described in Sect.~\ref{model_prep}.
Once inference is completed, we have to parse the raw YOLO output detection into final detection, consisting of bounding box, object class and class score. At first, we determine a class of an object by choosing the one with the highest score for every detection in raw output. Next, the detections are filtered by a class score. The threshold can be selected using a precision-recall curve and depends on the requirements of the target application. Elements with scores lower than the given threshold are rejected. The next step is to perform \textit{Non-Maximum Suppression} (NMS)~\cite{nms} to discard overlapping boxes and select the best one. 


The inference step of neural network inference produces a 2D bounding box on the image. We use the time of the original image acquisition from the camera-to-world matrix to project this 2D bounding box into the 3D scene observed by the user in AR. 
Based on this implementation, we can adequately annotate the 3D position of the object even if the user is looking in a different direction than when we captured the image for object detection.

\section{Evaluation}
The proposed processing pipeline with YOLOv8 is evaluated under two key criteria for the final application: (1) processing time and (2) object detection performance. All experiments were performed using the \textit{val2017} subset of Microsoft COCO dataset~\cite{coco}. This image data collection is a large-scale object detection, segmentation, and captioning dataset containing 91 categories of ordinary objects. It is a common choice in object detection tasks regarding benchmarking methods and using COCO pretrained models to perform transfer learning and fine-tuning to adapt models to different detection tasks \cite{yolotransfer}. Weights of YOLOv8 models pretrained on COCO are available online. 

We used \verb|System.Diagnostics.Stopwatch| for processing time measurement with high-resolution performance counter mode on capturing either the whole processing or a selected part of the pipeline. 
Each time measurement was repeated $100$ times after the model warm-up, i.e., several initial inferences that are necessary for each GPU application to stabilize the processing times and further ensure fair and robust comparisons between different configurations. 
The HL2 battery charge level was over $50\%$ for all trials, and the headset was not connected to any other device or power source.

\subsection{Measuring the impact of YOLOv8 model size}
The first design choice we have to make when using the YOLOv8 detector is selecting the network's size among the available family of architectures. 
The YOLOv8 authors publicly share five pretrained networks that can be used out-of-the-box in the desired application, starting from the smallest network to the more extensive networks measured as a number of parameters: (i) nano (YOLOv8n), (ii) small (YOLOv8s), (iii) medium (YOLOv8m), (iv) large (YOLOv8l), and (v) extra large (YOLOv8x). Selecting a smaller network hinders the final performance while simultaneously taking less memory on the device and providing faster inference. 
Simultaneously, a larger network offers a better object detection performance.

\begin{figure}[!b]
    \centering
    \includegraphics[width=\columnwidth]{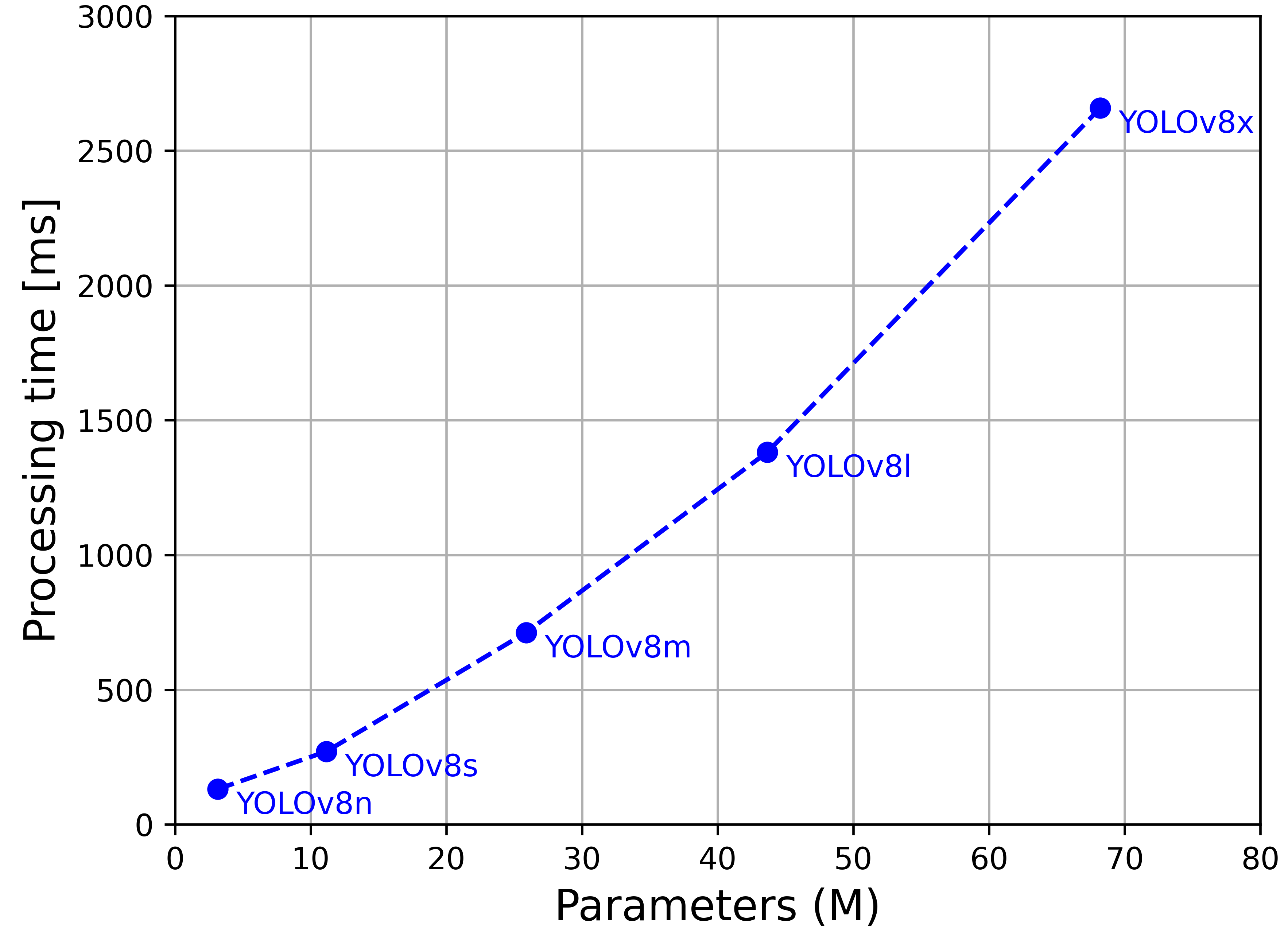}
    \caption{The single-image, onboard HL2 processing times depending on the chosen YOLOv8 model size for a fixed image size $224\times224$ pixels.}
    \label{fig:model_variant_vs_processing_time}
\end{figure}

Our experiments measured the processing time from the image capture moment to the bounding boxes projected in the 3D view. 
The comparison was performed with a usual image size of $224\times224$ pixels, and the results are presented in Fig.~\ref{fig:model_variant_vs_processing_time}.
These results suggest that for the best user experience in dynamic scenes, only the smallest YOLOv8n can meet the real-time requirements. 
Other models can still work in the resource-constrained environment of HL2. 
However, they might only be suited for scenarios where real-time performance is not vital for user experience, e.g., when used to classify the object held in hand or when the scene is not dynamic.
In these cases, the object detection will still be able to properly place the object position in the user's surroundings but will require more time to get these results. 

\begin{figure}[h!]
    \centering
    \includegraphics[width=\columnwidth]{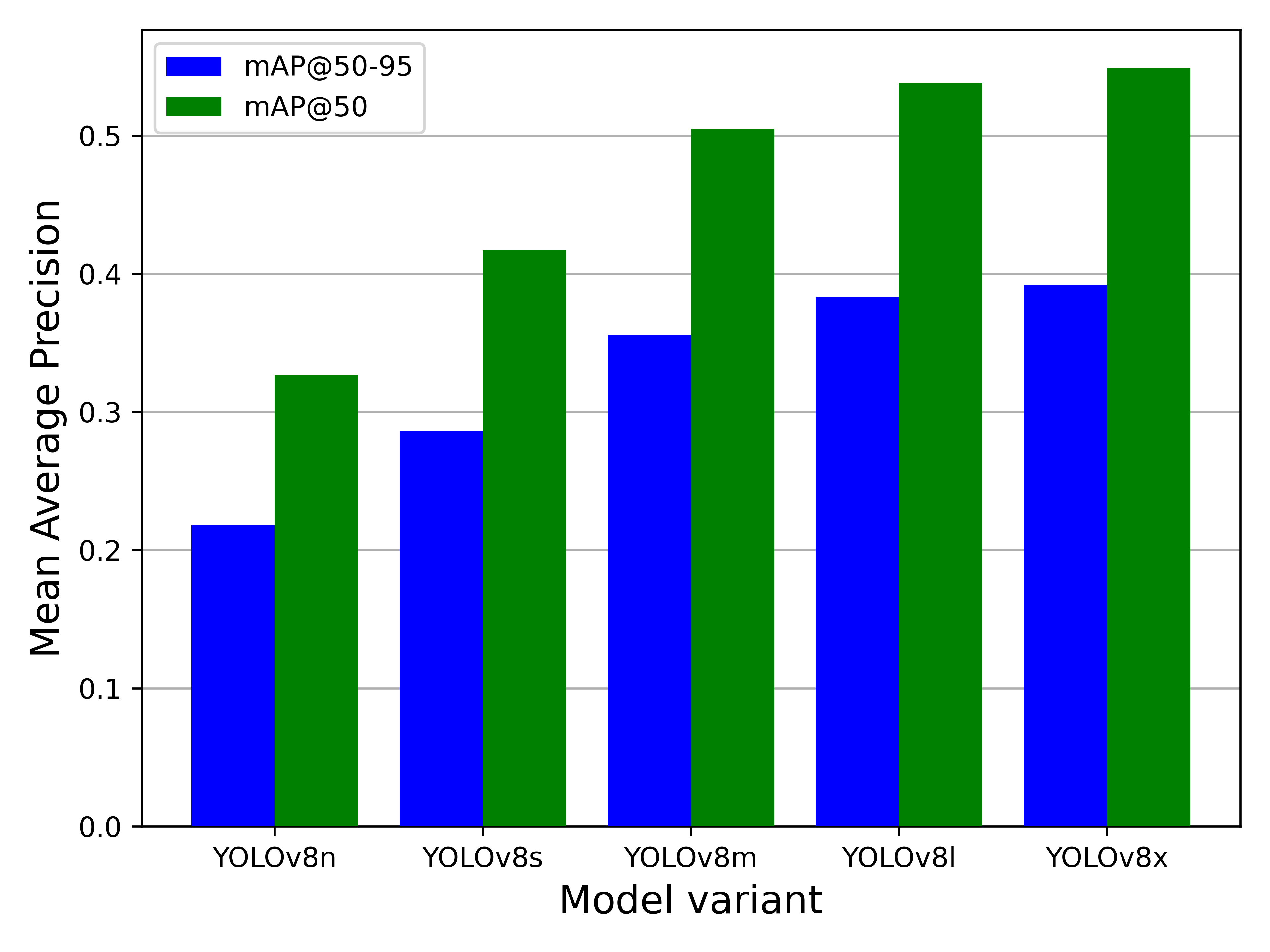}
    \caption{The measured performance of object detection as mAP values depending on the network size for YOLOv8 starting from the smallest network (YOLOv8n) to the largest network (YOLOv8x) for a fixed image size $224\times224$ pixels.}
    \label{fig:coco_map_vs_model_size}
    \vspace{-3mm}
\end{figure}

The usage of network models with a lower number of parameters results in lower performance. The usual metric to quantify performance is mean average precision (mAP), which is the average precision for all object classes measured at a selected threshold \verb|A|. 
\verb|mAP@A| indicates the performance when at least a \verb|A|$\%$ overlap between the bounding box from detection and ground truth bounding box (Intersection over Union -- IoU) is required to assume that the object was correctly recognized. 
The performance of different detection model configurations is presented in Fig.~\ref{fig:coco_map_vs_model_size} for \verb|mAP@50| and \verb|mAP@50-95| averaging the performance over a range of IoU thresholds. 
The obtained results suggest that a significant drop in performance should be expected when using smaller models.

\subsection{Object detection depending on the input image size}

Unfortunately, even with the smallest model, i.e., YOLOv8n, we ought to seek further improvements to achieve real-time performance dictated by the best immersive experience for AR headset users.

\begin{figure}[!hb]
    \centering
    \includegraphics[width=\columnwidth]{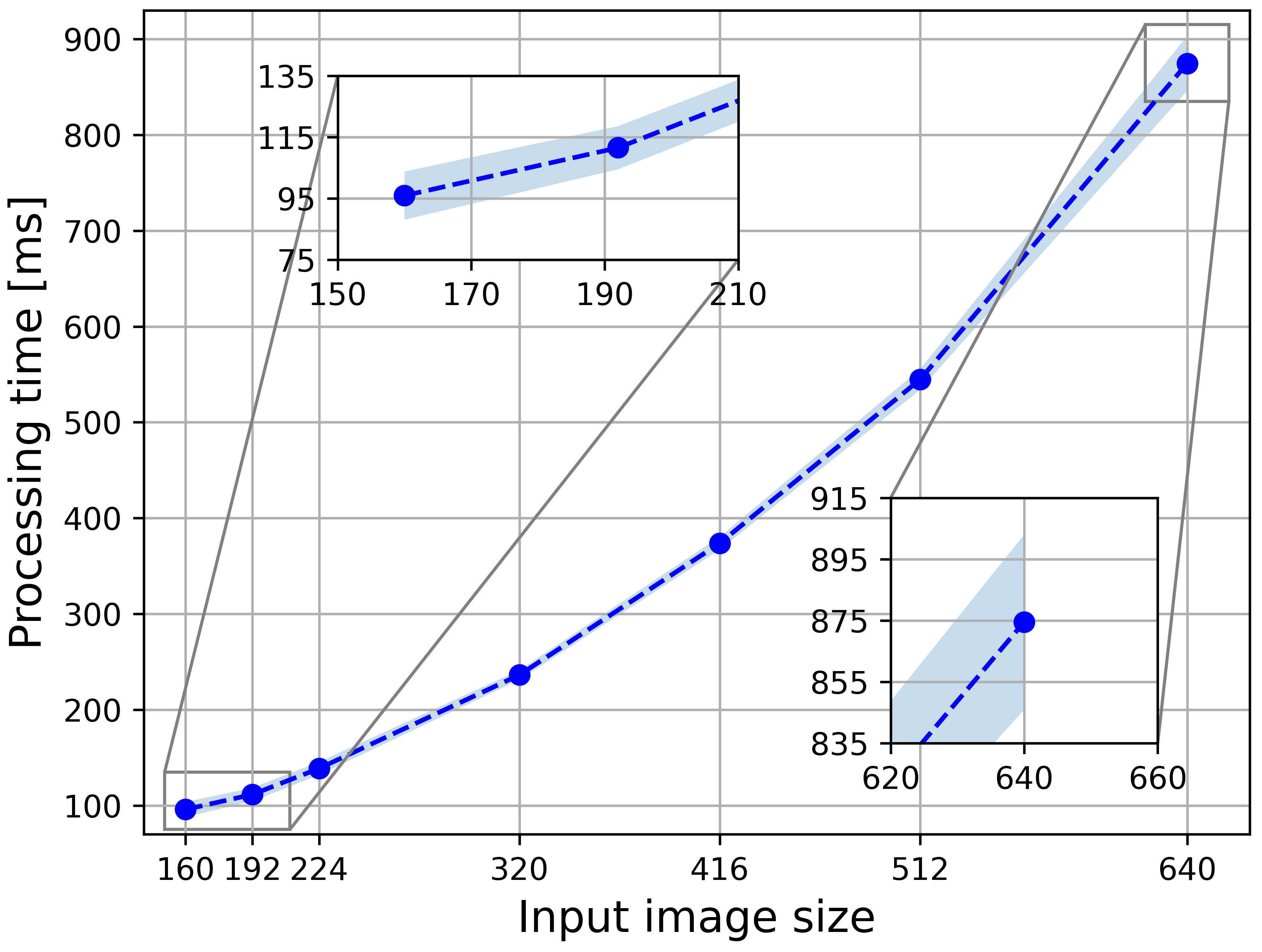}
    \caption{The total processing time for object detection for YOLOv8n model with different input image sizes. The light blue interval shows the standard deviation of the performed measurements.}
    \label{fig:input_size_vs_processing_time}
\end{figure}

\begin{figure}[!th]
    \centering
    \includegraphics[width=\columnwidth]{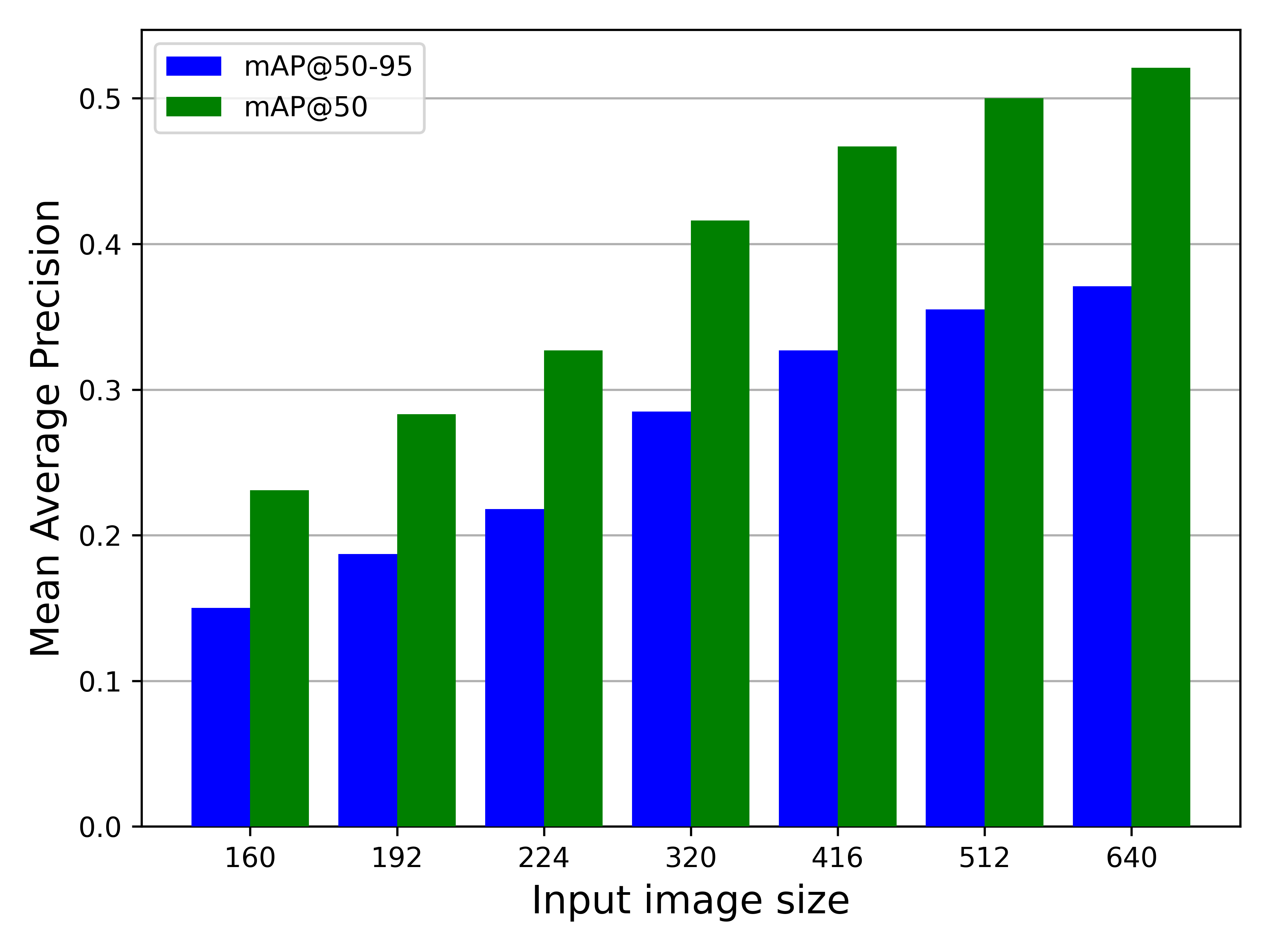}
    \caption{The measured performance of object detection using YOLOv8n model with different input image sizes.}
    \label{fig:coco_map_vs_input_size}
\end{figure}

Apart from the size of the network, the other possibility is to reduce the input image size as it directly impacts the inference times.
The results we obtained for varying image input sizes are presented in Fig.~\ref{fig:input_size_vs_processing_time}. 

The obtained relation between processing times and an input image size shows that processing times scale almost quadratically with the side length of an image (i.e. linearly with the number of pixels). 
Based on this observation, we can see that it is possible to obtain object detection results in less than 100 [ms] when using an image size of $160 \times 160 $ pixels.
Using smaller input image sizes might impact the achieved performance of the algorithm. We show the influence of the input image size on the mAP of the algorithms in Fig.~\ref{fig:coco_map_vs_input_size}.

\subsection{Choosing the best model based on processing time budget}
We might also have a greater processing time budget depending on the application. In these scenarios, we wanted to quantify if using a larger network for inference is more beneficial or, rather, increasing the backbone size to improve the network's detection performance makes sense.

\begin{figure}[htbp!]
    \centering
    \includegraphics[width=\columnwidth]{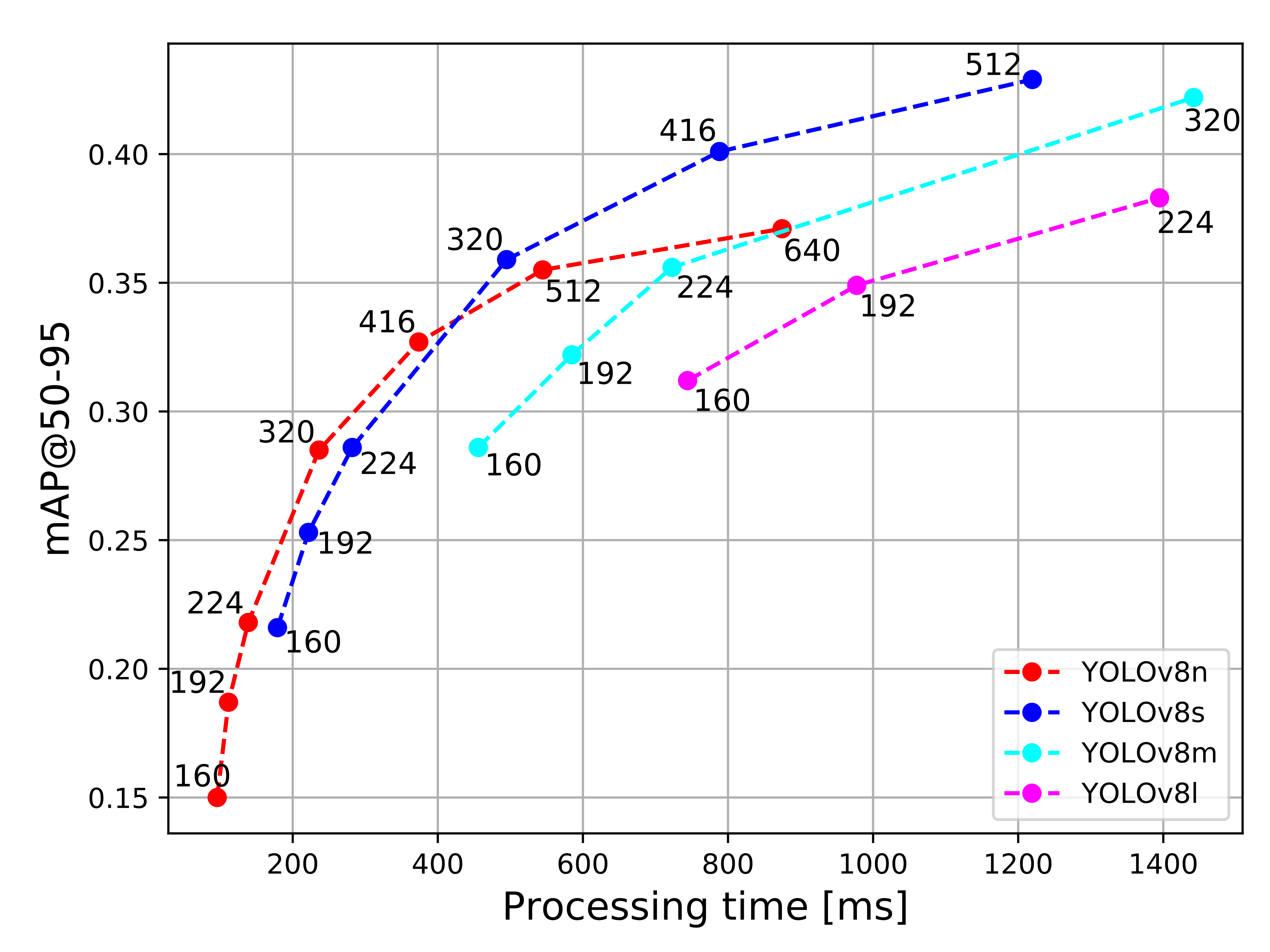}
    \caption{Comparison of mAP and inference time for different sizes of YOLOv8 models}
    \label{fig:model_map_vs_time_variants}
\end{figure}

The obtained results suggest that for any application with an inference time budget below 400 [ms], it is beneficial to use YOLOv8n while tuning the image size to fit the budget requirements. 
Compared to YOLOv8n at the same processing time, larger networks perform worse as they need to use smaller images.
For processing times thresholds greater than 400 [ms], we should choose YOLOv8s as it offers better performance than YOLOv8n despite smaller input image sizes than YOLOv8n while outperforming all larger backbones in the analyzed processing time interval. 
The presented conclusions are drawn based on the obtained performance on all objects in the COCO dataset, which might not hold equally for particular object classes.

\subsection{Model performance analysis for AR applications}


\begin{figure*}[htbp!]
     \centering
     \begin{subfigure}[b]{0.18\textwidth}
         \centering
         \includegraphics[width=\textwidth]{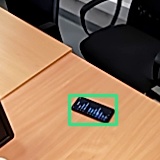}
         \caption{}
         \label{fig:dist_1}
     \end{subfigure}
     \begin{subfigure}[b]{0.18\textwidth}
         \centering
         \includegraphics[width=\textwidth]{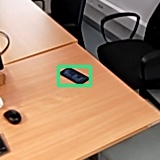}
         \caption{}
         \label{fig:dist_1_5}
     \end{subfigure}
     \begin{subfigure}[b]{0.18\textwidth}
         \centering
         \includegraphics[width=\textwidth]{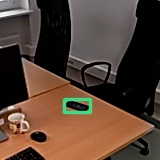}
         \caption{}
         \label{fig:dist_2}
     \end{subfigure}
      \begin{subfigure}[b]{0.18\textwidth}
         \centering
         \includegraphics[width=\textwidth]{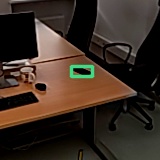}
         \caption{}
         \label{fig:dist_2_5}
     \end{subfigure}
      \begin{subfigure}[b]{0.18\textwidth}
         \centering
         \includegraphics[width=\textwidth]{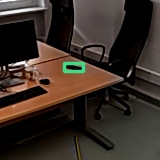}
         \caption{}
         \label{fig:dist_3}
     \end{subfigure}
        \caption{The performance of the proposed real-time YOLOv8n when using $160 \times 160$ input image size. The network detects objects reliably from (a) 1~[m], (b) 1.5~[m], (c) 2.0~[m], (d) and 2.5~[m] distance, with decaying results up to (e) 3.0~[m].}
        \label{fig:yolo_use_cases}
        \vspace{-3mm}
\end{figure*}

Flustered by the reduced performance of the YOLOv8n with a small input image size of $160 \times 160$, we conducted a series of real-world test experiments.
We focused on an example object, i.e. a smartphone, detected from 1 [m] up to 4 [m] with an object observed from 20 different viewing angles at each distance as presented in Fig.~\ref{fig:yolo_use_cases}. 
We selected these distances, arguing that reaching up to 2 [m] is crucial for AR interaction due to the maximum extent of human arms and hand-held tools. We frequently encounter such situations when dealing with shop floor tasks \cite{Tadeja2022ARQ} or device repair \cite{Bozzi2023Print3D}, which require close vicinity, i.e., arms-stretch distance from non-digital asset users are interacting with.

\begin{figure}[htbp!]
	\centering
	    \includegraphics[width=0.99\columnwidth]{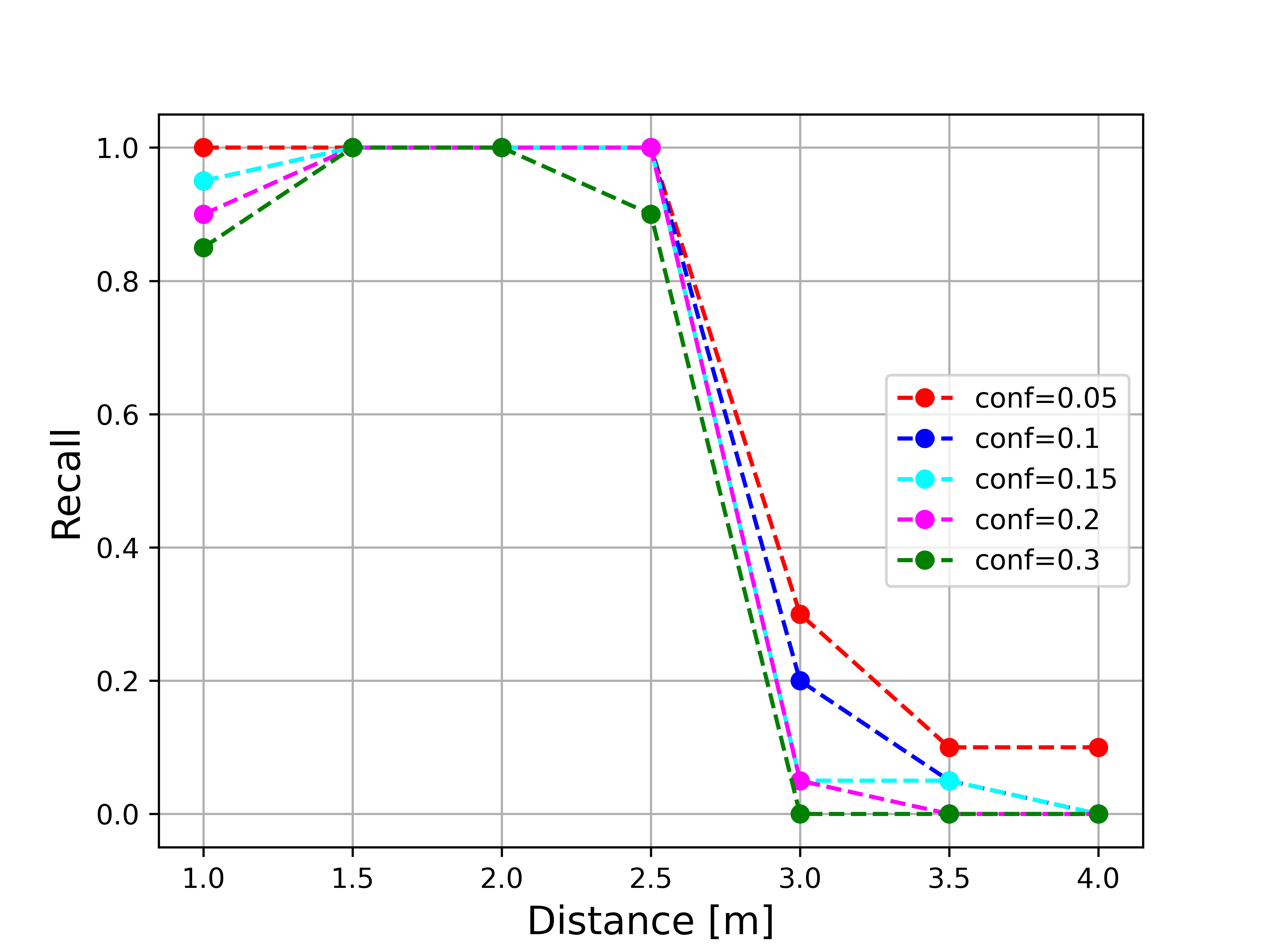}
	\caption{Recall depending on the distance to the smartphone for different confidence thresholds based on the real-time YOLOv8n run on $160\times160$ input image sizes. The obtained results suggest that the proposed real-time configuration of YOLOv8n on HL2 can be sufficient for most AR applications.}
	\label{fig:cellphone_recall}
 \vspace{-4mm}
\end{figure}

We measured the performance at each distance as a recall, i.e., a ratio of the number of cases when the smartphone was properly detected to the number of images. 
The results obtained for different confidence thresholds are presented in Fig.~\ref{fig:cellphone_recall}.
These outcomes suggest that the proposed configuration offering inference results in less than 100 [ms] can still detect all object (i.e. smartphone) instances if we focus on distances closer than 2.5 [m]. We believe that such performance fulfills the requirements for most AR use case scenarios \cite{Tadeja2022ARQ, Bozzi2023Print3D}.

\section{Ablation study}

The goal of the following section is to understand further the limitations of the YOLOv8 with $160 \times 160$ input image size and propose a solution that could further speed up the processing capabilities depending on the application.

\subsection{Processing time analysis}

As a first step, we measured the time spent to prepare the data (i.e., preprocessing), make the actual bounding box prediction (i.e., inference), and the time necessary to analyze the obtained predictions (i.e., postprocessing).
The results are summarized in Tab.~\ref{tab:best_yolo} and indicate the most time spent performing the inference. 
It shows that postprocessing, even though the NMS step is not performed on the GPU, is not a limiting factor. 

\begingroup
\setlength{\tabcolsep}{10pt}
\renewcommand{\arraystretch}{1.5}
\begin{table}[!ht]
\centering
\begin{tabular}{l|cc}
                        & \multicolumn{2}{c}{\textbf{processing time}}    \\ \hline
\textbf{operation}      & \textbf{mean {[}ms{]}} & \textbf{std {[}ms{]}} \\ \hline
\textbf{preprocessing}  & 1.97                   & 1.49                  \\ \hline
\textbf{inference}      & 89.84                  & 7.45                  \\ \hline
\textbf{postprocessing} & 4.06                   & 1.17                  \\ 
\end{tabular}
\caption{The analysis of the total processing time when using the YOLOv8n model and $160 \times 160$ images size input revealing that the most time is spent doing the core neural model inference}
\label{tab:best_yolo}
\end{table}
\endgroup

Further analysis of the inference time using the Unity profiler tool revealed that the time spent to copy the data for inference is negligible, taking less than $1\%$ of the overall inference time. 
Therefore, further improvements should be sought in the inference itself.

\subsection{Testing different model processing backends}

The processing time analysis indicates that we should improve the inference time. One possibility is to explore the inference backends available in the Barracuda package \cite{barracuda_backends} 
The backend choice determines whether the neural network will be run on GPU or CPU and what kind of implementation will be used. 
The results received for different backends are presented in Tab.~\ref{tab:backends}.

\begingroup
\setlength{\tabcolsep}{10pt}
\renewcommand{\arraystretch}{1.5}
\begin{table}[!ht]
\centering
\begin{tabular}{c|ccc}
\multicolumn{1}{l}{} & \multicolumn{1}{l}{} & \multicolumn{2}{c}{\textbf{inference time}} \\ \hline
\multicolumn{1}{l}{\textbf{device}} & \textbf{backend} & \textbf{mean {[}ms{]}} & \textbf{std {[}ms{]}} \\ \hline
\multirow{3}{*}{\textbf{GPU}} & \textbf{Compute} & 103.64 & 6.68 \\ \cline{2-4} 
 & \textbf{ComputeRef} & 174.65 & 6.65 \\ \cline{2-4} 
 & \textbf{\begin{tabular}[c]{@{}c@{}}Compute\\ Precompiled\end{tabular}} & 89.84 & 7.45 \\ \hline
\multirow{2}{*}{\textbf{CPU}} & \textbf{CSharp} & 344.83 & 23.31 \\ \cline{2-4} 
 & \textbf{CSharpBurst} & 216.11 & 25.62 \\
\end{tabular}
\caption{Inference times for different backend selections on HL2 using YOLOv8n and image input size of $160\times160$}
\label{tab:backends}
\end{table}
\endgroup

The fastest inference times were obtained for \verb|ComputePrecompiled| backend, followed by the \verb|Compute| backend, which did not improve further results. We were unable to execute an inference with remaining \verb|CSharpRef| and \verb|PixelShader| backends. 

Another way to improve the neural network inference performance is the usage of quantization, either with reduced float precision (FP16) or integer (INT8).
Unfortunately, the Barracuda library does not support the FP16 nor INT8 quantization in its current implementation. 
FP16 usually offers a significant speed-up compared to the full float implementations as already proven on other computing platforms~\cite{fp16,fp16_app}.

\subsection{Dealing with non-square image input}
One commonly used approach to dealing with larger images and smaller objects assumes sliding the object detector over the whole image~\cite{sliding_window}.
Consequently, we wanted to verify if it is beneficial for HL2 to divide the input image into smaller sub-images while increasing the batch size of the passed data input.
Here, another gain is that the view from HL2 is not square, as we have a greater horizontal field of view than the vertical one.
Our experiment compared two approaches: the network with $320\times320$ input image size and a network input of $320\times160$ pixels divided into two sub-images, each of size $160\times160$ pixels, passed as a batch size two for network inference.
The results are summarized in Tab.~\ref{tab:batch_size}.

\begingroup
\setlength{\tabcolsep}{10pt}
\renewcommand{\arraystretch}{1.5}
\begin{table}[!ht]
\centering
\begin{tabular}{l|cc}
 & \multicolumn{2}{c}{\textbf{detection time}} \\ \hline
\textbf{image size} & \textbf{mean {[}ms{]}} & \textbf{std {[}ms{]}} \\ \hline
\textbf{2 x 160 x 160} & 133.64 & 3.94 \\ \hline
\textbf{1 x 320 x 320} & 236.31 & 4.98 \\ 
\end{tabular}
\caption{Inference time comparison between the YOLOv8  processing the $320\times320$ input image size compared to the two side-stacked images of $160\times160$ input images passed as batch size two.}
\label{tab:batch_size}
\vspace{-4mm}
\end{table}
\endgroup

The total processing time for two side-stacked images is almost half the time required to process the square-size single image. 
These results confirm that the inference time linearly depends on the number of image pixels and that the proposed batch approach allows the process of non-square image inputs to preserve the natural aspect sizes of objects. 
Retraining the model for non-square inputs would require more effort than the proposed sliding window approach.

\section{Conclusions}
This paper presents the steps required for real-time object detection using the state-of-the-art YOLOv8 network model on the Microsoft HoloLens 2 edge computing platform. 

We believe that the ability to run advanced ML such as YOLOv8 algorithms directly on an HMD will become necessary for the emerging edge-based virtual reality applications \cite{edgemetaverse} and the \textit{Metaverse} concept \cite{acm2023surv}. For a wide range of practical applications, particularly educational \cite{Igras2023VRSpeecRecon} and medical \cite{Zikas2023Med}, the VR and AR technologies need to converge \cite{Li2023}, providing the users with an ability to seamlessly interact with both real and virtual elements of the surroundings.

To that end, our experiments show the universal path that can be taken to ensure real-time operation by finding the best trade-off between the neural network model size and the input image size. In the presented case of an AR headset, we are forced to reduce the input image sizes to $ 160 \times 160 $ pixels, still obtaining satisfactory results from the perspective of AR applications mainly within the 2.5~[m] range for an object of interest detection. Such distance span is well-aligned with the typical operational range of AR use case scenarios \cite{Syberfeldt2016TypesOfAR, Tadeja2022ARQ}.

The analysis of processing times (see Tab. \ref{tab:best_yolo}) reveals that further improvements should be sought in the inference itself, as preprocessing, postprocessing, or data copying take little time. Beyond that, to boost the number of possible use cases of the proposed solution, we also show that slicing a view into multiple images processed in a single batch can be a sensible approach when we are dealing with situations where wider FoV is used.

In our future work, we plan to tackle the missing FP16 support to reduce inference times further. In addition, we will also incorporate tracking into object detection to extend the possible applications of the presented framework.


\end{document}